\documentclass{article}

\usepackage{arxiv}

\usepackage[utf8]{inputenc} 
\usepackage[T1]{fontenc}    
\usepackage{hyperref}       
\usepackage{url}            
\usepackage{booktabs}       
\usepackage{amsfonts}       
\usepackage{nicefrac}       
\usepackage{microtype}      
\usepackage{lipsum}		
\usepackage{graphicx}
\usepackage{cite}

\usepackage{doi}

\usepackage{caption}
\usepackage{subcaption}

\title{Continuation of Famous Art with AI:\\A Conditional Adversarial Network Inpainting Approach}

\author{ \href{https://orcid.org/0000-0002-9858-1231}{\includegraphics[scale=0.06]{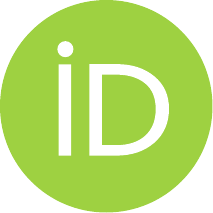}\hspace{1mm}Jordan J.~Bird}\thanks{\url{https://jordanjamesbird.com/}} \\
	Computational Intelligence and Applications Research Group (CIA)\\Department of Computer Science\\Nottingham Trent University\\Nottingham, United Kingdom\\
	\texttt{jordan.bird@ntu.ac.uk} \\
}

\renewcommand{\headeright}{Jordan J. Bird}
\renewcommand{\shorttitle}{Continuation of Famous Art with AI}

\hypersetup{
pdftitle={Continuation of Famous Art with AI: A Conditional Adversarial Network Inpainting Approach},
pdfauthor={Jordan J. Bird},
pdfkeywords={Image-to-Image Translation, Conditional Generative Adversarial Networks, Computational Creativity, AI Art}
}

\begin{document}
\maketitle

\begin{abstract}
Much of the state-of-the-art in image synthesis inspired by real artwork are either entirely generative by filtered random noise or inspired by the transfer of style. This work explores the application of image inpainting to continue famous artworks and produce generative art with a Conditional GAN. During the training stage of the process, the borders of images are cropped, leaving only the centre. An inpainting GAN is then tasked with learning to reconstruct the original image from the centre crop by way of minimising both adversarial and absolute difference losses, which are analysed by both their Fr\'echet Inception Distances and manual observations which are presented. Once the network is trained, images are then resized rather than cropped and presented as input to the generator. Following the learning process, the generator then creates new images by continuing from the edges of the original piece. Three experiments are performed with datasets of 4766 landscape paintings (impressionism and romanticism), 1167 Ukiyo-e works from the Japanese Edo period, and 4968 abstract artworks. Results show that geometry and texture (including canvas and paint) as well as scenery such as sky, clouds, water, land (including hills and mountains), grass, and flowers are implemented by the generator when extending real artworks. In the Ukiyo-e experiments, it was observed that features such as written text were generated even in cases where the original image did not have any, due to the presence of an unpainted border within the input image.
\end{abstract}


\section{Introduction}
\begin{figure}
    \centering
    \includegraphics[scale=1]{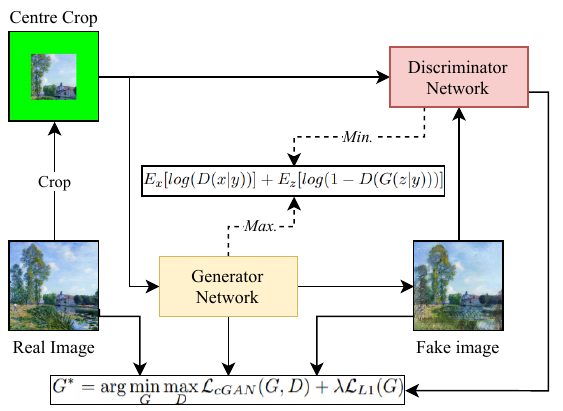}
    \caption{A general overview of the training approach followed by this work, with adversarial training and L1 loss reduction used to reconstruct images from a small section of the original.}
    \label{fig:general-overview}
\end{figure}
First and foremost, Computational Creativity is collaborative work between the fields of artificial intelligence, psychology, philosophy, and the arts, which endeavour to explore whether machines are capable of human-level creativity. If Ada Lovelace's objection to machine intelligence is considered, then Computational Creativity still remains an endeavour to achieve an understanding and enhancement of creativity without the machine being creative in the same sense that a human being can be. Regardless of whether human imagination can be replicated or not, contemporary research and industry are beginning to explore the concept of synthetic media, wherein data is artificially modified, manipulated, and produced. To name just a few benefits of media synthesis, modern entertainment enjoys the generation of subtitles, visual effects, music, palette selection, post-production, and more. 

\begin{figure}
\centering
\begin{subfigure}{.5\textwidth}
  \centering
  \includegraphics[width=.7\linewidth]{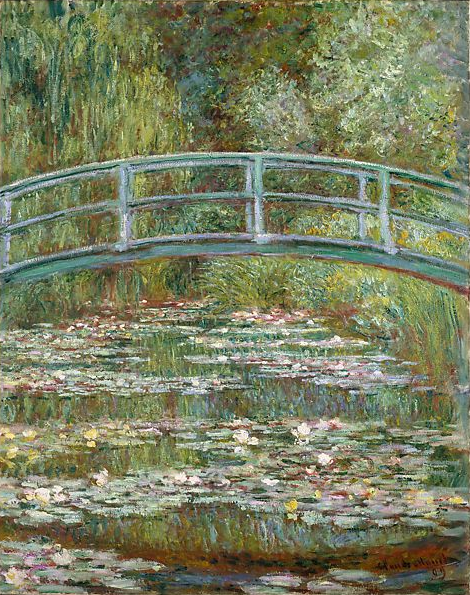}
  \caption{Original work.}
  \label{fig:monet-sub}
\end{subfigure}%
\begin{subfigure}{.5\textwidth}
  \centering
  \includegraphics[width=.7\linewidth]{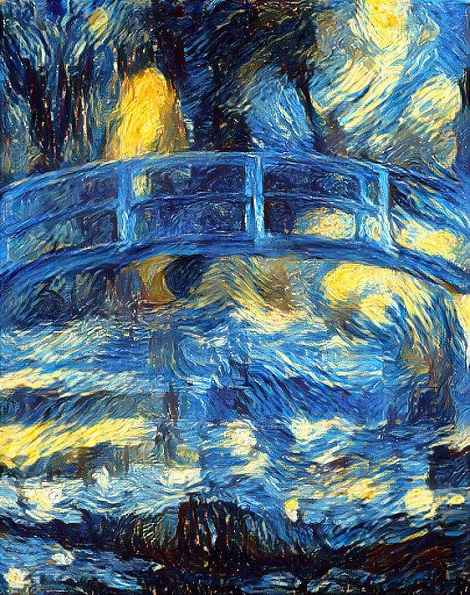}
  \caption{Work with style transfer applied.}
  \label{fig:monet-style}
\end{subfigure}
\caption{The Water Lily Pond by Claude Monet (1899) with style transferred from Starry Night by Vincent van Gogh (1889).}
\label{fig:test}
\end{figure}

Paintings find their limits at the edge of the canvas. For example, in 1899, Monet painted the famous \textit{Water Lily Pond}, depicting his water-lily garden with a wooden bridge crossing the pond. As can be seen in Figure \ref{fig:monet-sub} the framing of the work ends within the midst of the trees and to the edges of the bridge, begging questions to our visual imagination such as, \textit{``what did the sky look like?"} or \textit{``how far did the pond stretch?"}. Typically, techniques to generate AI art focus on two main approaches. First and foremost generation of new works in their entirety, e.g. algorithms may learn from the eighteen paintings in Monet's series of the pond of water lilies and then generate a nineteenth based on the learning process and generalisation. The second common approach is to perform style transfer, in relation to this example, style transfer would allow for the same work to be re-imagined in the style of another, such as Van Gogh's \textit{Starry Night} which can be observed in Figure \ref{fig:monet-style}.

The approach presented in this work differs from the state of the art given that it explores the possibility of continuing and extending famous artworks through an image inpainting approach. To refer to the questions posed earlier in this section, the ability of inpainting with a focus towards the outer limits of the canvas would allow for the unsupervised generation of new work which treats a full painting as a starting point. Regarding Monet's painting, the generalisation ability of a landscape model may learn that forests and trees do not continue forever, instead meeting the sky at a canopy of treetops. Indeed, if the process is also repeated several times, then the centre of the painting (the original work) would become increasingly smaller before disappearing completely, leaving behind only the generative output in the form of an entirely new image. 

The two main research questions that this work will explore are (i) \textit{``how far can an artwork be reconstructed by means of only a section of the image?"} and, if the approach is possible, \textit{``how can reconstruction of artwork be used to continue famous works and create new images?"}. The scientific contributions of this work surround the application of conditional GAN architectures to create generative images by extending famous artworks. The benefits and the limitations of the approach are explored, and models plus weights are shared for future work and applications by other researchers. 

Following this introduction, Section \ref{sec:background} explores the background of the field and reviews related work. Following this, Section \ref{sec:method} explains the method behind this approach. Results are presented in Section \ref{sec:results} and observations are made by exploring the behaviours and possibilities of the model. Finally, this work is concluded and future work is suggested in Section \ref{sec:conclusion}.

\section{Background}
\label{sec:background}
The synthesis of new media by machines far outdates computers, historians noting that the first known synthetic forms of media date from 60AD~\cite{brett1954automata,sharkey2007programmable}. Ancient Greek Automata such as those designed by Daedalus and Hero of Alexandria physically wrote text and played music. Inevitably, this captivation led to more recent approaches to synthesising media with algorithms that attempt to model the creative process, a prominent example of this is Harold Cohen's AARON which began in 1973~\cite{mccorduck1991aaron}. AARON is directed by hard-coding styles of art, before the algorithm can then generate an infinite number of images in the programmed style.

Many debates surrounding artwork generated by Artificial Intelligence focus on who (or what) gets the credit for the generated output. As noted by Epstein et al.~\cite{epstein2020gets}, there is much variation to how members of the public perceive AI, ranging from just computers and electricity to a more anthropomorphic perception. This study also noted that perception is easily manipulated by the choice of the language used to describe machines and algorithms. Indeed, the relevance of this debate has evolved from the realm of thought experiment to that which has serious financial implications; thanks, in large part, to the auction of an artwork generated by a Deep Convolutional Generative Adversarial Network (DCGAN) at Christie's. Pre-auction estimates ranged from \$7,000 to \$10,000, but the work instead sold for \$432,500~\cite{schneider2018has}. Herein lies two issues that are currently debated in the state of the art - Firstly, the code for the model was written by Robbie Barrat who was not affiliated with the project, nor received royalties~\cite{caldas2020generative}. Secondly, if a GAN is trained on the work of a thousand artists, does the programmer deserve the credit for the generated outputs or are they instead a product of those one thousand artists? John Dewey argued that knowledge is socially constructed and based on past experiences~\cite{dewey1986experience}, providing arguments for humans to create work based on their past experiences and imagination without plagiarising their muse. To provide such an argument for machines on the other hand leads to a great many unanswered philosophical questions on the notions of sentience and machine intelligence - an argument made famous by Ada Lovelace's objection and Alan Turing's responses~\cite{abramson2008turing}. Plagiarism is indeed possible by GANs as noted in ~\cite{gayadhankar2021image}, where generative models were observed to fool modern plagiarism detection approaches. 

A growing number of modern image synthesis with a focus on artworks are based on Radford et al.'s Deep Convolutional Generative Adversarial Network (DCGAN)~\cite{radford2015unsupervised}. Although the research remains unpublished as a preprint, it has retrieved over ten thousand citations at the time of writing with published applications including the generation of artworks~\cite{nasrin2020hennagan,pradhan2020implementation,xue2021end}, text-to-image synthesis~\cite{cha2017adversarial,cho2020design}, converting greyscale images by colourisation~\cite{suarez2017infrared,blanch2019end}, and data augmentation~\cite{frid2018gan,zhu2021data}. In their 2017 paper, Tan et al.~\cite{tan2017artgan} suggested an extension to DCGAN similar to that of the CGAN, noting that backpropagation regarding categorical labels could improve the generation of artworks. In 2020, Hertzmann~\cite{hertzmann2020visual} described the GAN's visually indeterminant behaviour to bare likeness to that of modern representational art. A prominent example of this is Pepperell's Succulus~\cite{wang2020toward}. Visual indeterminacy is a byproduct of data generalisation, and causes outputs to resemble reality yet defy spatial interpretation on examination, where familiar objects are only suggestive~\cite{ishai2007perception,wallraven2007eye}. Hertzmann noted that BigGan~\cite{brock2018large} was particularly likely to produce more indeterminate images. In \cite{elgammal2017can}, the authors show that by using affective information, images can be generated by a GAN that would ultimately be noted as artworks by human observers. Affective language itself was also considered by \cite{achlioptas2021artemis}, where a deep learning model was able to learn nuanced language to describe feelings evoked by a painting by learning from human descriptions. Image inpainting is the term used to describe the use of computational techniques to fill-in missing parts of an image while aiming to remain natural and undetected~\cite{bertalmio2000image}. In~\cite{xie2012image}, the authors proposed that images could be effectively denoised through inpainting via a \textit{denoising auto-encoder}. This included both random static noise as well as images which were overlayed with text (which was removed). The work also noted that there was the ability to transfer learn between different types of noise, for example, a network trained to denoise gaussian noise could denoise salt-and-pepper noise at 90.05\% accuracy. Image inpainting is often used to reconstruct damaged images, such as digital scans of old photographs that have been creased due to folding~\cite{richard2001fast}. Reconstruction of both form and texture are two very different problems, but recent advances in computational intelligence have shown that the two problems can be solved simultaneously~\cite{bertalmio2003simultaneous,pnevmatikakis2008inpainting,ren2019structureflow}. Structural knowledge has also shown to alleviate this problem~\cite{yang2020learning}. 

Following much human-led direction, the idea of the Generative Adversarial Network (GAN) was presented~\cite{goodfellow2020generative}. Originally introduced in 2014 and published in 2020, the GAN is composed of two neural networks; a Generator and a Discriminator. The two networks learn to generate data and discriminate between real and generated data, respectively. This approach leads to a zero-sum game competition, wherein the loss of one of the networks is beneficial to the other. The output of the generator network feeds directly into the discriminator network and thus learning is automated through competition. Loss can be calculated as follows:
\begin{equation}
\label{eq:gan}
    L_{GAN}(G,D) = E_{x}[log(D(x))] + E_{z}[log(1-D(G(z)))],
\end{equation}
where $E_{x}[log(D(x))]$ is the classification of real images and the second part $E_{z}[log(1-D(G(z)))]$ is the classification of generator outputs (i.e., recognition that the image is fake). $E_{x}$ and $E_{z}$ are the expected values over all real and fake data, e.g., $x$ is a real image from the dataset and $z$ may be a random noise vector which is grown and filtered through convolutions which have the aim of generating a new data object. $D(x)$ is the probability of an input being real, the reverse being the probability of the input being synthetically created by the generator. This is known as a \textit{minimax loss}, since the generator's aim is to maximise Equation \ref{eq:gan} while the discriminator aims to minimise it.
To extend upon the above example, the Conditional Generative Adversarial Network (CGAN or Conditional GAN)~~\cite{mirza2014conditional} allows inputs to be conditioned. For example, with a class label. Equation \ref{eq:gan} is therefore extended as follows:
\begin{equation}
\label{eq:cgan}
    L_{cGAN}(G,D) = E_{x}[log(D(x|y))] + E_{z}[log(1-D(G(z|y)))],
\end{equation}
where data objects $x$ and $D(z)$ are conditioned on class label $y$. Therefore, $D(x|y)$ is the discriminator's probability that $x$ is real given class label $y$, and $G(z|y)$ is the output of the generator with random vector $z$ given class label $y$. 
For image-to-image translation, the task that this work faces, Isola et al. suggest two main extensions to the classical GAN approach~\cite{isola2017image}. Firstly, the implementation of L1 loss, which is the minimisation of the sum of the absolute difference between generator outputs and ground truth data. L1 loss in this regard is formulated as:
\begin{equation}
    \label{eq:l1}
    L_{1}(G) = E_{x,z}[| y - G (x,z) |].
\end{equation}
\begin{figure}
    \centering
    \includegraphics[scale=0.65]{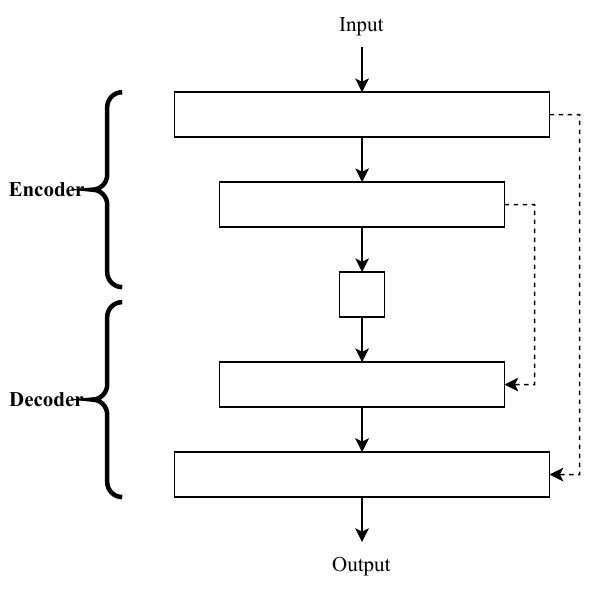}
    \caption{Diagram of a U-Net architecture, residual connections exist between the encoder and decoder rather than a linear relationship.}
    \label{fig:unet}
\end{figure}
Secondly, the paper also suggests the use of a U-Net architecture~\cite{ronneberger2015u}. As can be observed in Figure \ref{fig:unet}, the U-Net architecture differs from an encoder-decoder network by implementing skip connections or residual information~\cite{he2016deep}. Rather than solely downsampling to a bottleneck before upsampling to an output, the reasoning suggested for the use of a U-Net is to directly transfer the low-level information between earlier and later layers. To give a specific example related to the problem in this work, certain textures and entities such as tree bark and leaves may hold useful information prior to their downsampling and so to transfer them directly by concatenation to the decoder may aid in improving the quality of the painting. 

The approach presented in this work differs from the current state of the art due to the following considerations: (i) The supervised approach to the dataset for inpainting provides a method to generate the borders of an image by learning to undo a centre crop. (ii) Following the supervised approach to learning, rather than generating entirely new works, the model instead takes famous artworks and generates new borders following a shrink operation rather than the crop required for supervised learning. Therefore, ultimately, the trained model is able to take a full artwork and continue it from the outer edges of the image. In terms of the state-of-the-art, this is the first time that inpainting has been explored as a method to continue artworks.

\section{Method}
\label{sec:method}
The general idea behind the approach can be observed in Figure \ref{fig:general-overview}. During the training process, 512px images are taken and the centre 256$\times$256 pixels are cropped. After being removed, the borders of the image are replaced with a bright green colour inspired by ``chroma-key" (R:G:B 0:255:0). This colour is chosen to differentiate the missing borders from the image, i.e., to avoid situations where a black border may be incorporated into darker artwork. Regarding the then-cropped centre, the original image is treated as the ground truth for model training. Following this training process, given positive progress on the image inpainting problem, the model can then generate new artwork by providing input as a resized image rather than a centre crop. 
\begin{figure}
    \centering
    \includegraphics{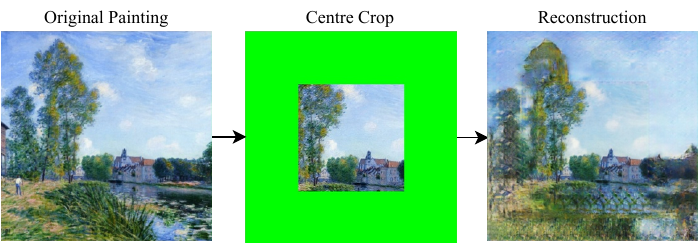}
    \caption{Visual example of model training; a centre crop of the image is taken and the goal of the generator is to reconstruct the original image.}
    \label{fig:training-model}
\end{figure}
\begin{figure}
    \centering
    \includegraphics{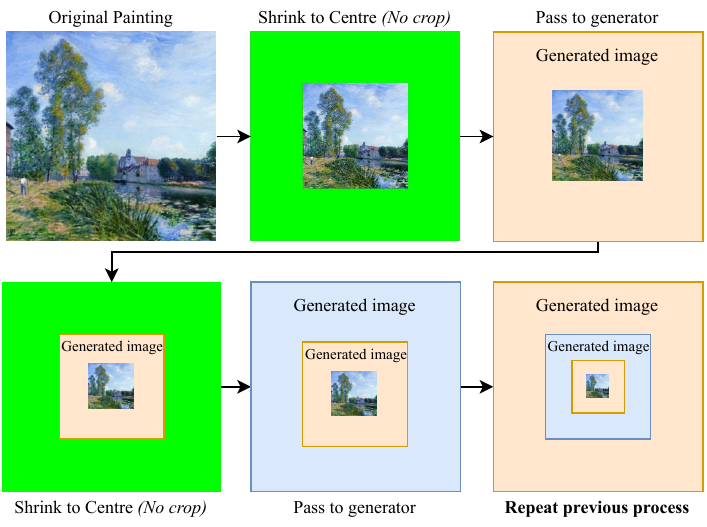}
    \caption{Visual example of image generation. The trained model is tasked with inpainting areas of the image that do not exist in reality (given the image is shrunk and not cropped). This process, when repeated, leads the original image to disappear after several iterations.}
    \label{fig:generating-images}
\end{figure}
To give a visual example of these explanations, Figure \ref{fig:training-model} shows how the model is trained in a supervised fashion; the artwork is cropped and the generator's goal is to reconstruct the original image. In Figure \ref{fig:training-model}, the trained model is then given the same task, except the input image is downsized rather than cropped, and so the output are reconstructed parts of the image that do not actually exist in reality. The process described by Figure \ref{generating-images} can be used to repeat the process and generate more and more data, while shrinking the original image to the centre. This process can be repeated as many times as required. 
\begin{table}[]
\centering
\caption{Number of images within the three datasets used in the experiments.}
\label{tab:dataset-details}
\footnotesize
\begin{tabular}{@{}llll@{}}
\toprule
                            \textbf{Genre }  & \textit{\textbf{Landscape }} & \textit{\textbf{Ukiyo-e }} & \textit{\textbf{Abstract }} \\ \midrule
\textbf{Num. Images } & 4766                & 1167             & 4968              \\ \bottomrule
\end{tabular}
\end{table}
Table \ref{tab:dataset-details} details the sizes of the datasets used in these experiments. The images are downloaded from Wikiart~\cite{wikiartwebsite}, and low resolution images are manually removed after physical inspection. Images are cropped and resized to 512 pixel squares for uniform input. Given the findings of the DCGAN paper~\cite{radford2015unsupervised}, image pixels are normalised to the range $[-1,1]$. 

In addition to this finding, the paper also suggests that convolutional layers have a LeakyRELU activation function which enables small negative values and aids in the avoidance of mode collapse. The generator contains 8 downsampling layers, the first having 64 filters, the second having 128, the third 256, and the remaining downsampling layers have 512 filters. The upsampling part of the U-Net has seven layers, the first four containing 512 filters, the fifth having 256, the sixth having 128, and the final upsampling layer has 64 filters. The discriminator network is a PatchGAN, introduced by Demir and Unal in 2018~\cite{demir2018patch}. The downsampling layers in the PatchGAN contain 64, 128, and 256 layers, respectively. Optimisers for both networks are ADAM with a learning rate of $2e-4$ and a beta 1 decay of 0.5. The network is trained with a batch size of 1. Fr\'echet Inception Distance (FID)\cite{frechet1957distance,heusel2017gans}, calculated as:
\begin{equation}
FID =\left\|\mu-\mu_{w}\right\|_{2}^{2} + tr \left(\Sigma+\Sigma_{w}-2\left(\Sigma \Sigma_{w}\right)^{1 / 2}\right),
\end{equation}
is finally used to evaluate the models at each 10 epoch interval. The FID is measured between the generated images and the expected output, using the InceptionV3 model\cite{szegedy2016rethinking}. 

The models were implemented with TensorFlow and trained via a GPU, on the 4352 CUDA cores of an Nvidia RTX 2080Ti. Each model was trained for 150 epochs in total. 

\section{Results and Observations}
\begin{table}[]
\centering
\caption{Fr\'echet Inception Distance at 10 epoch intervals for each of the datasets during training.}
\label{tab:FID-table}
\begin{tabular}{@{}rrrr@{}}
\toprule
\multicolumn{1}{l}{\textbf{Epoch}} & \multicolumn{1}{l}{\textbf{Landscapes}} & \multicolumn{1}{l}{\textbf{Ukiyo-e}} & \multicolumn{1}{l}{\textbf{Abstract}} \\ \midrule
\textit{\textbf{10}}               & 5.120                                   & 17.095                               & 18.929                                \\
\textit{\textbf{20}}               & 4.491                                   & 16.732                               & 17.053                                \\
\textit{\textbf{30}}               & 3.884                                   & 14.306                               & 15.819                                \\
\textit{\textbf{40}}               & 3.457                                   & 13.940                               & 14.640                                \\
\textit{\textbf{50}}               & 2.919                                   & 12.440                               & 13.527                                \\
\textit{\textbf{60}}               & 3.114                                   & 10.936                               & 12.563                                \\
\textit{\textbf{70}}               & 3.411                                   & 9.860                                & 11.812                                \\
\textit{\textbf{80}}               & 3.439                                   & 9.549                                & 11.320                                \\
\textit{\textbf{90}}               & 3.546                                   & 8.982                                & 11.255                                \\
\textit{\textbf{100}}              & 3.299                                   & 8.417                                & 10.334                                \\
\textit{\textbf{110}}              & 3.551                                   & 8.263                                & 10.408                                \\
\textit{\textbf{120}}              & 3.642                                   & 7.967                                & 10.211                                \\
\textit{\textbf{130}}              & 3.412                                   & 7.653                                & 10.057                                \\
\textit{\textbf{140}}              & 3.414                                   & 7.564                                & 9.740                                 \\
\textit{\textbf{150}}              & 3.596                                   & 7.391                                & 9.399                                 \\ \bottomrule
\end{tabular}
\end{table}

\begin{figure}
    \centering
    \includegraphics[scale=0.65]{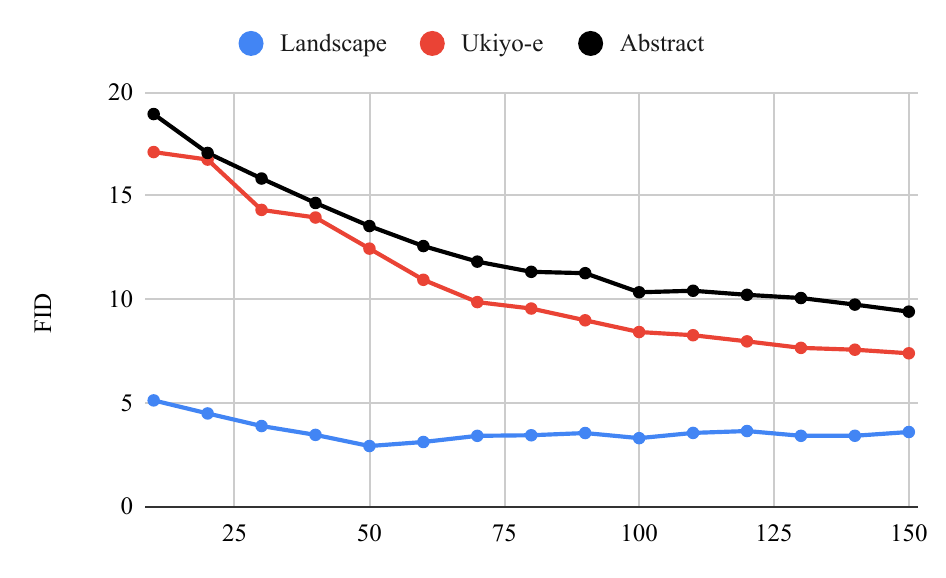}
    \caption{Fr\'echet Inception Distance (FID) at 10 epoch intervals for each of the datasets during training.}
    \label{fig:FID-graph}
\end{figure}

Table \ref{tab:FID-table} and Figure \ref{fig:FID-graph} show the Fr\'echet Inception Distance (FID) at each 10 epoch interval for the datasets. As can be noted, the Ukiyo-e and Abstract training processes reduce FID throughout. On the other hand, the landscape training finds a minimum FID at around epoch 50, with the distance increasing from that point onwards. 

\label{sec:results}
\subsection{Landscapes}
\begin{figure}
    \centering
    \includegraphics[scale=0.65]{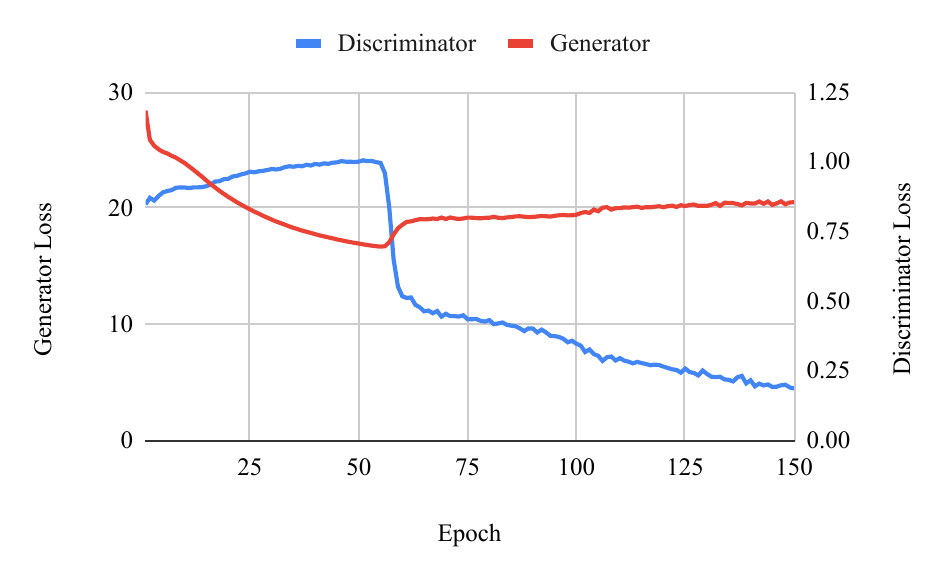}
    \caption{Observed Discriminator and Generator losses for the landscape art inpainting GAN.}
    \label{fig:landscape-disc-gen}
\end{figure}
\begin{figure}
    \centering
    \includegraphics[scale=0.65]{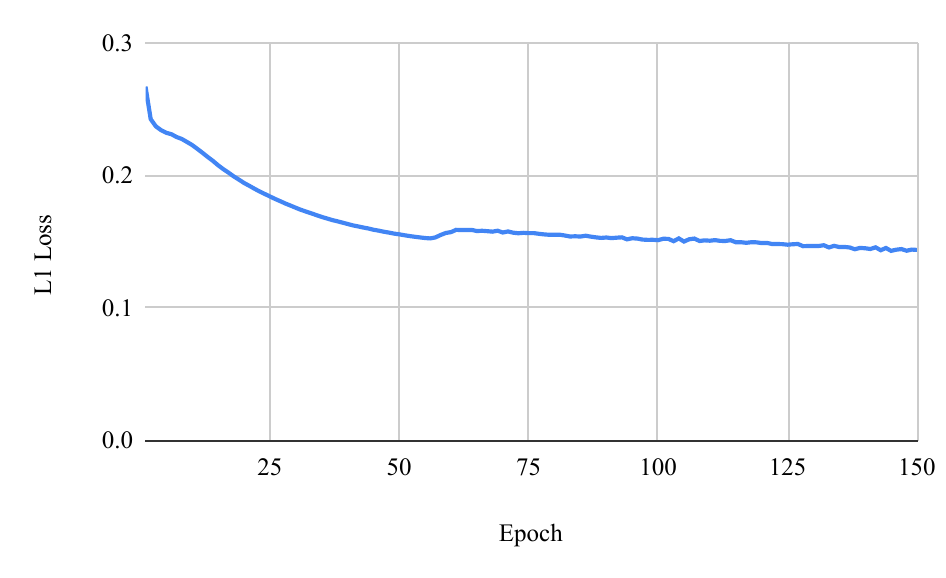}
    \caption{Observed generator L1 loss for landscape art inpainting.}
    \label{fig:landscape-l1}
\end{figure}
Figure \ref{fig:landscape-disc-gen} shows the discriminator and generator losses for the landscape experiment. It was noted that a drop in discriminator loss occurred, mode collapse on the other hand did not occur since learning continued. At epoch 57, discriminator loss was 0.834 before dropping to 0.647 at epoch 58. Figure \ref{fig:landscape-l1} shows the L1 loss or the landscape experiment, which gradually decreased throughout the process as expected. 
\begin{figure}
    \centering
    \includegraphics{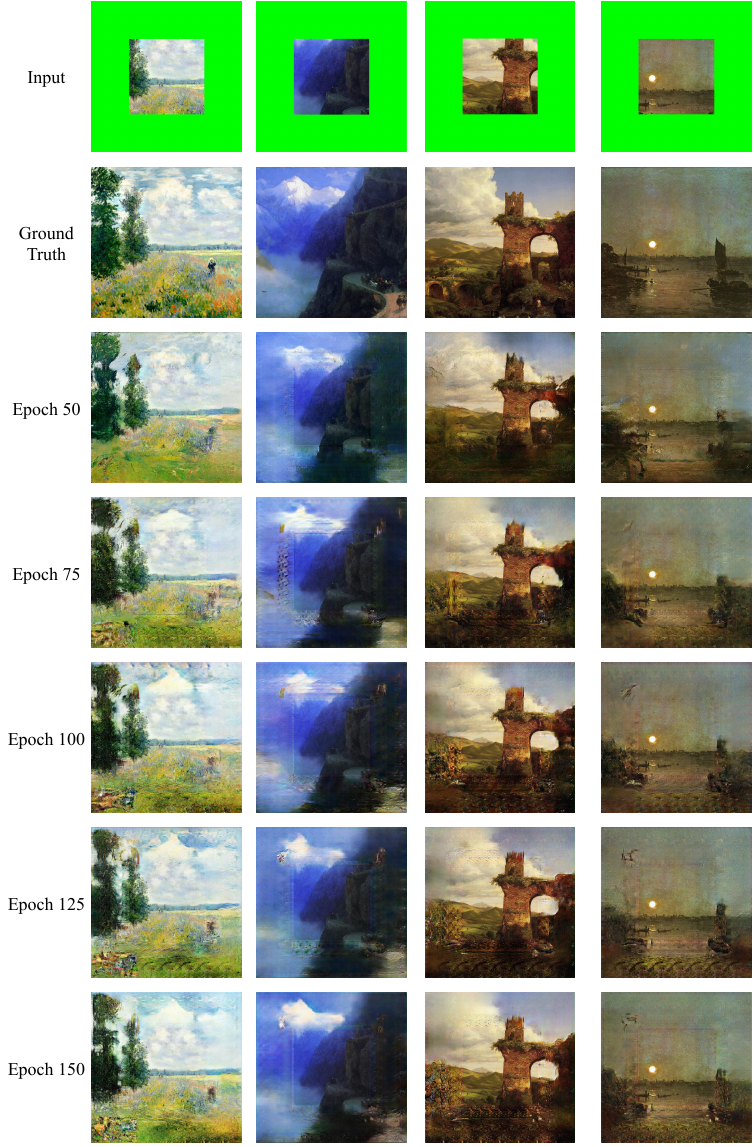}
    \caption{Examples of generator output for landscape art at selected epochs. }
    \label{fig:landscape-examples}
\end{figure}
Examples of training outputs can be observed in Figure \ref{fig:landscape-examples}. From left to right, the images shown for example are:
\begin{itemize}
    \item The Poppy Field near Argenteuil, Claude Monet (1873)
    \item Roads of Mljet to Gudauri, Ivan Aivazovsky (1868) 
    \item The Arch of Nero. Thomas Cole (1846)
    \item Moonlight, a Study at Millbank, Joseph Mallord William Turner (1797)
\end{itemize}
It was observed that fewer visual glitches were evident around epoch 50, but more detail could be observed alongside glitches later on. It can also be observed that finer features such as trees and the stone tower in particular gained proper form towards the later stages of the training process. 
\begin{figure}
    \centering
    \includegraphics[scale=0.7]{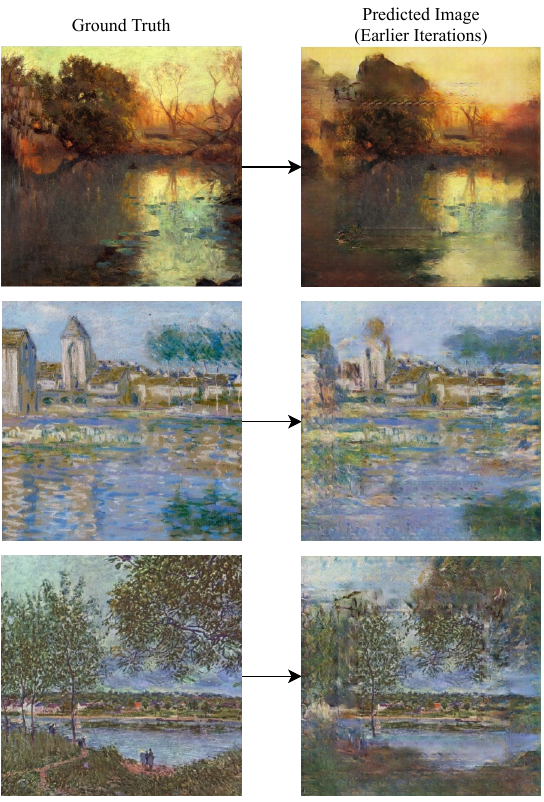}
    \caption{Observations at earlier iterations for landscape inpainting. During these iterations it was observed that a more generalised and blurred image is produced as a reflection of the original image.}
    \label{fig:landscape-blur}
\end{figure}
Figure \ref{fig:landscape-blur} shows some selected observations during the earlier epochs of the training process. It was noted that the inpainting does bare close similarity to the original image, but, during these stages, the generator was able to fool the discriminator by producing a less-detailed and blurrier image than should be expected. As can be seen in Figure \ref{fig:landscape-examples}, this was overcome during, and images with far more detail were produced later on. This observation provides an interesting insight to how the generator succeeds in reducing loss with this strategy of blurred generalisation.

\subsubsection{Continuation of landscape art}
\begin{figure}
    \centering
    \includegraphics{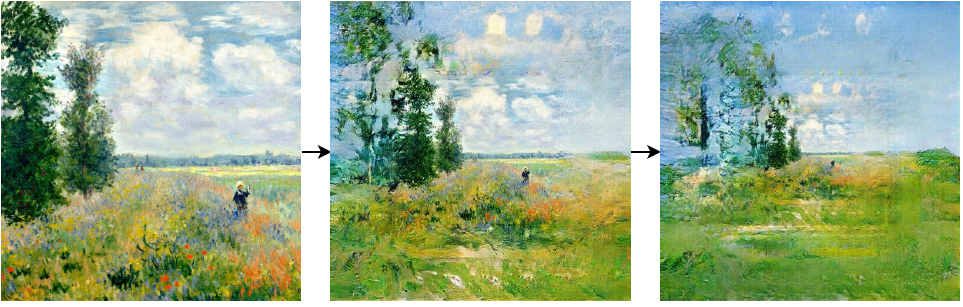}
    \caption{Two generations of continuation for The Poppy Field near Argenteuil, Claude Monet (1873).}
    \label{fig:landscape-continue-1}
\end{figure}
Figure \ref{fig:landscape-continue-1} shows two generations of continuation for The Poppy Field near Argenteuil by Claude Monet. In the first generation, we can note that the algorithm has caused the poppy field to disperse into a grassier area. The horizon in the distance is continued as well as the generation of new clouds. Interestingly, part of the poplar tree has been continued also, with what seems to be a third tree closer towards the point of view of the observer. In the second generation, the grassy area has become more general. A fourth tree has not been generated, with the colour of the sky painted in place. Further, more sparse clouds have been generated towards the top of the image.
\begin{figure}
    \centering
    \includegraphics{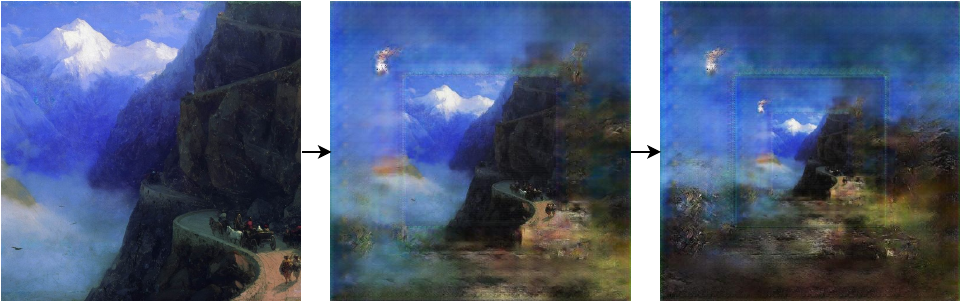}
    \caption{Two generations of continuation for Roads of Mljet to Gudauri, Ivan Aivazovsky (1868).}
    \label{fig:landscape-continue-2}
\end{figure}
Two generations continuation of Roads of Mljet to Gudauri by Ivan Aivazovsky can be observed in Figure \ref{fig:abstract-continue-2}. Interestingly, the generator seems to have some difficulty with this image in particular. It seems the fog in the distance has been considered an important feature, with visibility around the edges limited. This pattern continues to the second generation, though additional mountains and earth towards the point of view of the observer have been introduced.
\begin{figure}
    \centering
    \includegraphics{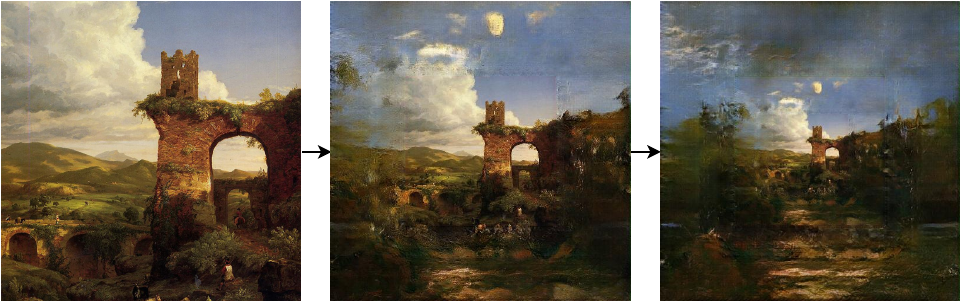}
    \caption{Two generations of continuation for The Arch of Nero. Thomas Cole (1846).}
    \label{fig:landscape-continue-3}
\end{figure}
In Figure \ref{fig:landscape-continue-3}, The Arch of Nero by Thomas Cole is continued for two generations. As the arch and watchtower become more distant and less distinct, natural features such as trees and cliffs are generated. Towards the closer areas from our point of view, what seems to be a path to the arch is generated. Throughout the generations, the sky is continued and naturally becomes a darker shade of blue at higher altitudes with further clouds generated.
\begin{figure}
    \centering
    \includegraphics{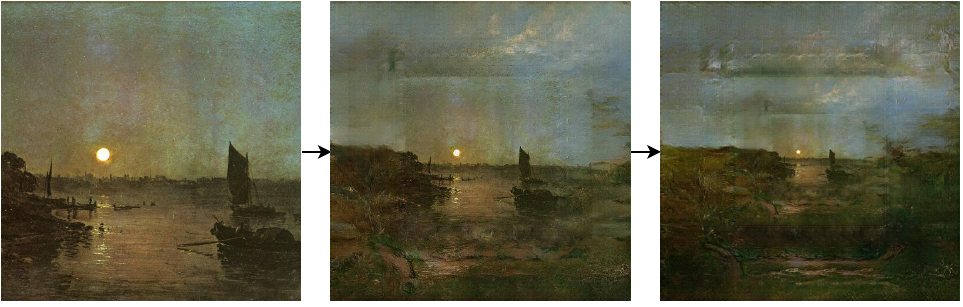}
    \caption{Two generations of continuation for Moonlight, a Study at Millbank, Joseph Mallord William Turner (1797).}
    \label{fig:landscape-continue-4}
\end{figure}
Figure \ref{fig:landscape-continue-4} shows the continuation of Joseph Mallord William Turner's Moonlight, a Study at Millbank. Similarly to that in Figure \ref{fig:landscape-continue-3}, the sky is continued and clouds are generated, and growing perspective causes the generation of natural features such as trees as well as a continuation of the water. 
\begin{figure}
    \centering
    \includegraphics{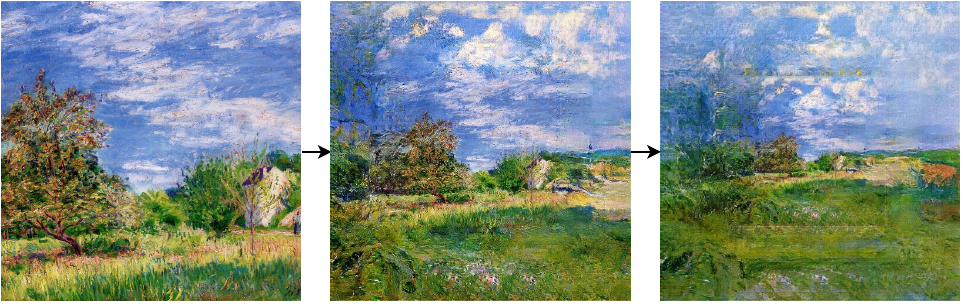}
    \caption{Two generations of continuation for Orchard in Spring, Alfred Sisley (1881).}
    \label{fig:landscape-continue-5}
\end{figure}
In the final experiment for landscape continuation, Figure \ref{fig:landscape-continue-5} shows the continuation of Orchard in Spring by Alfred Sisley. The continuation of the sky and clouds in both generations is particularly impressive, given that the clouds in the second image have arguably the same level of detail as in the original piece. It can be observed that the horizon is extended, with the yellowing grass seemingly transformed into a field when the perspective is taken into account, since the image is shrunk for each iteration. It can also be observed that white flowers, inspired by those in the original image, are produced within the grass when the generator extends the image and produces new data.

\subsection{Ukiyo-e}
\begin{figure}
    \centering
    \includegraphics[scale=0.65]{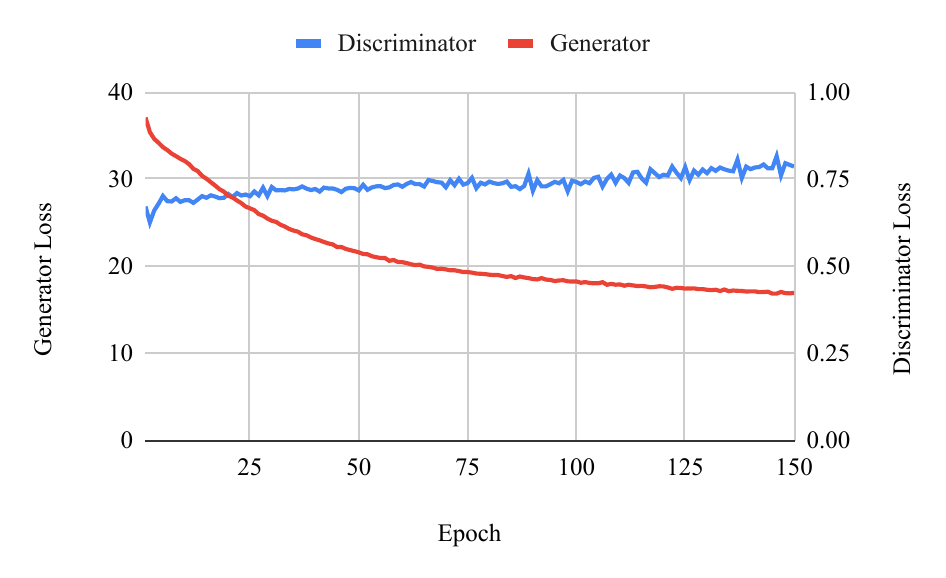}
    \caption{Observed Discriminator and Generator losses for the Ukiyo-e art inpainting GAN.}
    \label{fig:ukiyo-disc-gen}
\end{figure}
\begin{figure}
    \centering
    \includegraphics[scale=0.65]{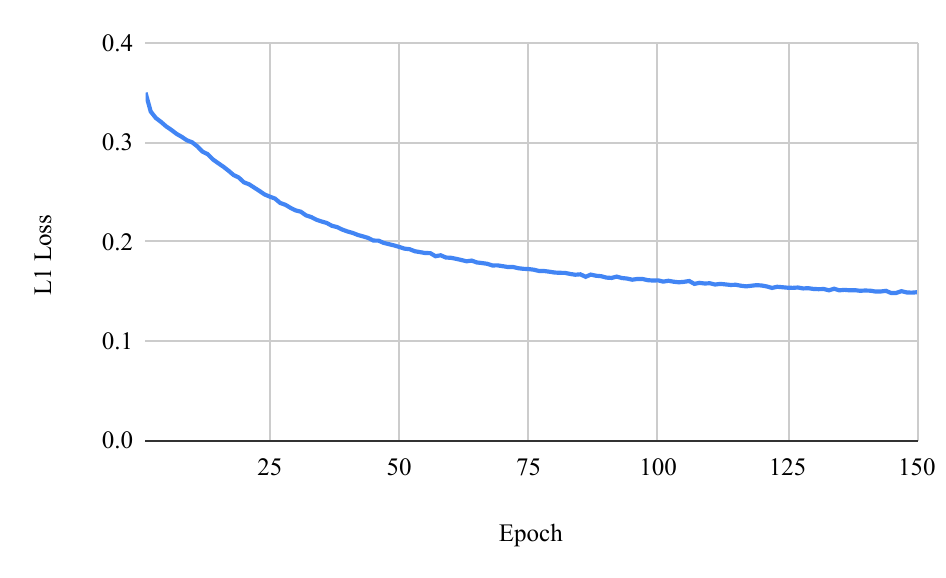}
    \caption{Observed generator L1 loss for Ukiyo-e art inpainting.}
    \label{fig:ukiyo-l1}
\end{figure}
Figures \ref{fig:ukiyo-disc-gen} and \ref{fig:ukiyo-l1} show the discriminator and generator losses, and the L1 loss for the Ukiyo-e experiment, respectively. As could be expected, the generator gradually increases in ability, being able to fool the discriminator to an increasingly positive extent. Towards the end of the training process, the generator's losses tend to become more stable and static relative to their previous values throughout.

\begin{figure}
    \centering
    \includegraphics{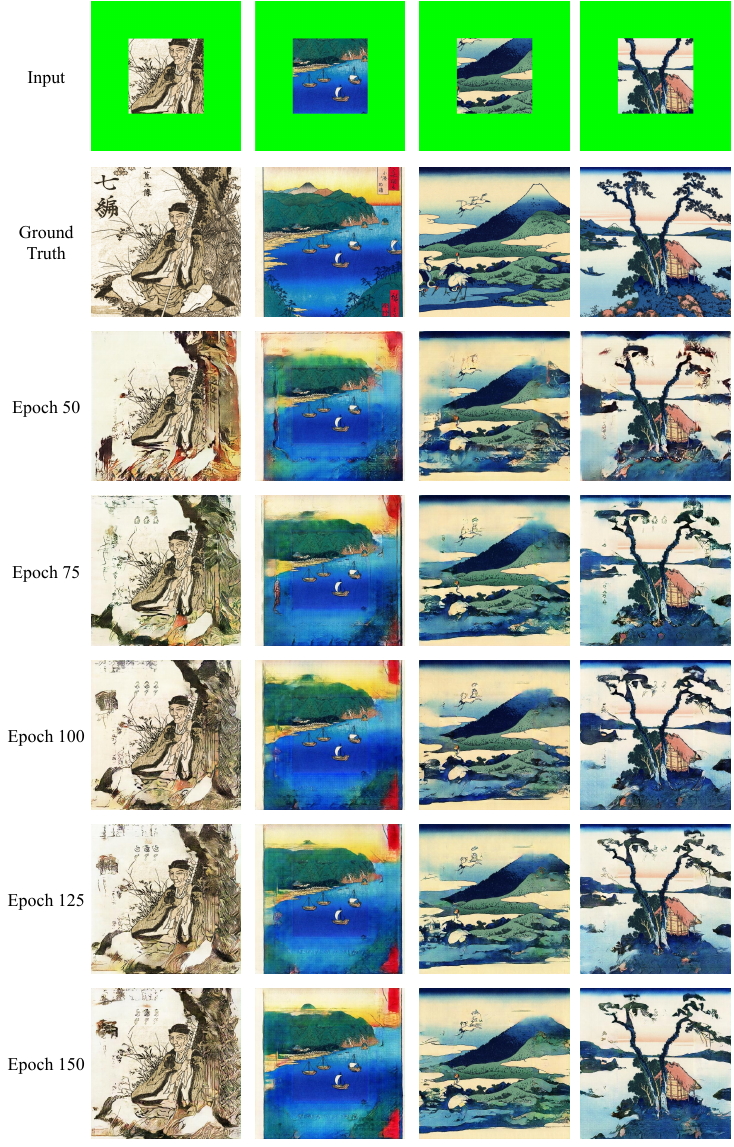}
    \caption{Examples of generator output for Ukiyo-e art at selected epochs. }
    \label{fig:ukiyo-examples}
\end{figure}
Examples of training outputs can be observed in Figure \ref{fig:ukiyo-examples}. From left to right, the images shown for example are:
\begin{itemize}
    \item Portrait of Matsuo Basho, Katsushika Hokusai (Edo period, 1603–1867) 
    \item Bay at Kominato in Awa Province, Utagawa Hiroshige (circa 1856)
    \item Umegawa in Sagami province, Katsushika Hokusai (circa 1830-1832) 
    \item Lake Suwa in the Shinano province, Katsushika Hokusai (circa 1829-1833) 
\end{itemize}
Similarly to the previous examples of generator outputs, it can be observed that finer features took more epochs to generate, as could be expected given the GAN learning process. Earlier outputs have an evident bounding box separating the input from the generated image, which fades throughout, implying that the discriminator began to recognise this box as a feature pointing towards an image being fake. It can be noted that style of the image was learnt by the generator. This can be seen when comparing the images at epoch 50, wherein the sketched image contains some red paint, which disappears almost entirely by epoch 100 and is replaced by a more realistic black ink. 

\begin{figure}
    \centering
    \includegraphics[scale=0.7]{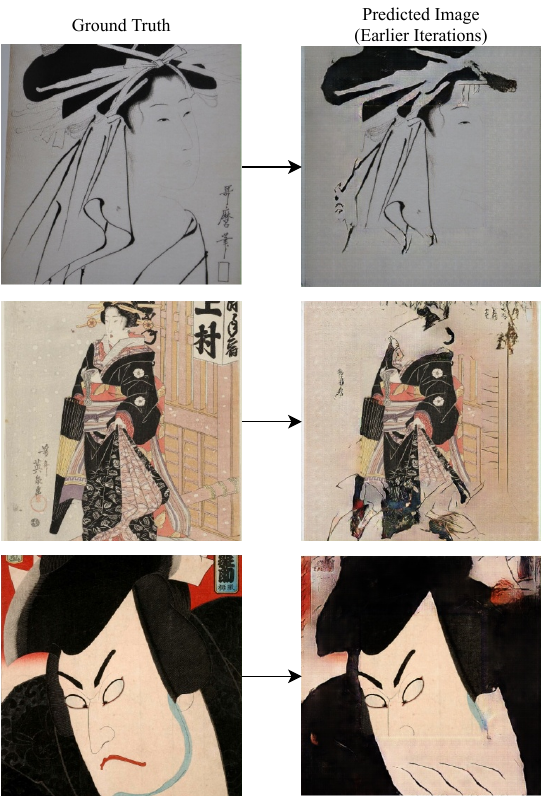}
    \caption{It was observed that the generator had particular difficulty when generating faces during the Ukiyo-e experiments. Note that realistic textures are produced, and the generator strategises to simply omit the faces.}
    \label{fig:ukiyo-nofaces}
\end{figure}
An interesting observation was made during training that can be observed within the examples in Figure \ref{fig:ukiyo-nofaces}. It was noted that although complex forms and shapes were generated, as can be seen, there was a particular difficulty faced when generating faces. Note that rather than making mistakes that would lead the discriminator to classify the image as fake, the generator instead strategises to produce realistic textures and simply omits the face. 

\subsubsection{Continuation of Ukiyo-e art}
\begin{figure}
    \centering
    \includegraphics{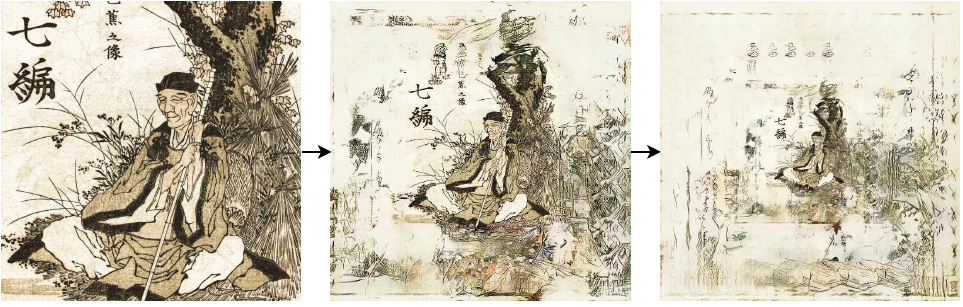}
    \caption{Two generations of continuation for the Portrait of Matsuo Basho, Katsushika Hokusai (Edo period, 1603–1867).}
    \label{fig:ukiyo-continue-1}
\end{figure}
Figure \ref{fig:ukiyo-continue-1} shows the generation of two continuations of the Portrait of Matsuo Basho by Katsushika Hokusai. Throughout both, the sketches of plant life around the portrait are continued towards the lower half of the image. Most interestingly, what seem to be shapes inspired by written Japanese characters appear throughout the image, continued from those that are present in the original work. The shapes towards the bottom of the image seem to resemble those of hills and mountains from works within the dataset, but they have been reproduced in the style of this image rather than their original form (usually they are painted in colour and have more detail). This is particularly interesting since it suggests that the model has been able to generalise to these forms but has also learnt to keep the stylisation of the input image. 
\begin{figure}
    \centering
    \includegraphics{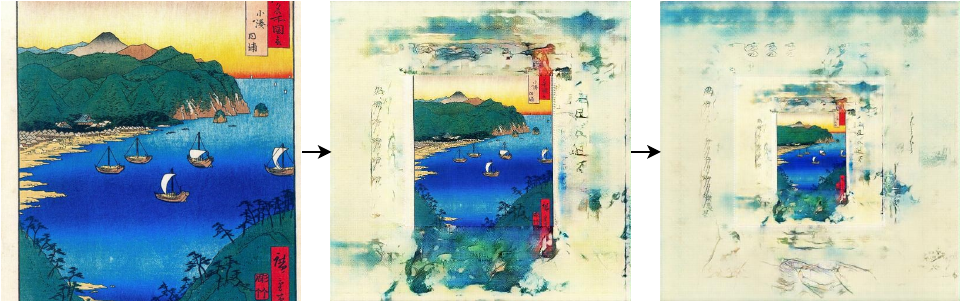}
    \caption{Two generations of continuation for the Bay at Kominato in Awa Province, Utagawa Hiroshige (circa 1856).}
    \label{fig:ukiyo-continue-2}
\end{figure}
The images in Figure \ref{fig:ukiyo-continue-2} show two generations of continuation of the Bay at Kominato in Awa Province by Utagawa Hiroshige. The interesting part in particular of these outputs are concerning the borders; the model has noted that there are borders to the left and right of the image and this is continued, with what seem to be written characters generated in a top-to-bottom sentence (as the model observed from the red banners within the image), similarly to those generated in the previous images within figure \ref{fig:ukiyo-continue-1}. Some confusion has occurred at the top and bottom of the image, which seems to be due to the nonexistence of these borders in those areas. Since this was not observed often, the paint patterns produced seem to bleed into the background in the style of watercolour. Towards the top right of the third image in the series, a shape which resembles a mountain has been generated, which is to be expected in some cases due to the prominence of Mount Fuji in many Japanese artworks~\cite{uhlenbeck2000mount,earhart2015mount} and thus the dataset. 
\begin{figure}
    \centering
    \includegraphics{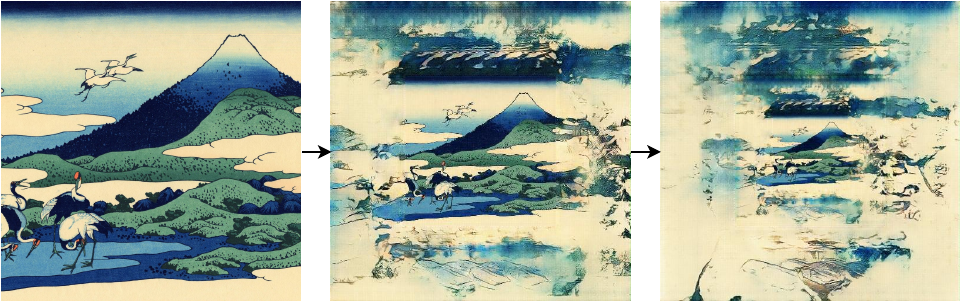}
    \caption{Two generations of continuation for Umegawa in Sagami province, Katsushika Hokusai (circa 1830-1832).}
    \label{fig:ukiyo-continue-3}
\end{figure}
Umegawa in Sagami province by Katsushika Hokusai is continued for two generations in Figure \ref{fig:ukiyo-continue-3}. Similarly to the previous example, the border at the top has caused some confusion and has been extended into a mountain range and clouds - there are no examples in the dataset of this gradient which represents the sky and edge of the image existing at a point in the image that is not the edge, thus, the generator must produce the rest of the image based on other features in the original work. Moreover, the aforementioned darkening blue gradient has been produced towards the top of the image after continuation. Towards the bottom of the final generation, a selection of shapes similar to a traditional boat seems to have been added to the image by the generator. The final thing to note is that, again similarly to previous experiments, written Japanese characters have been produced around the image, even though the original work has no text present.
\begin{figure}
    \centering
    \includegraphics{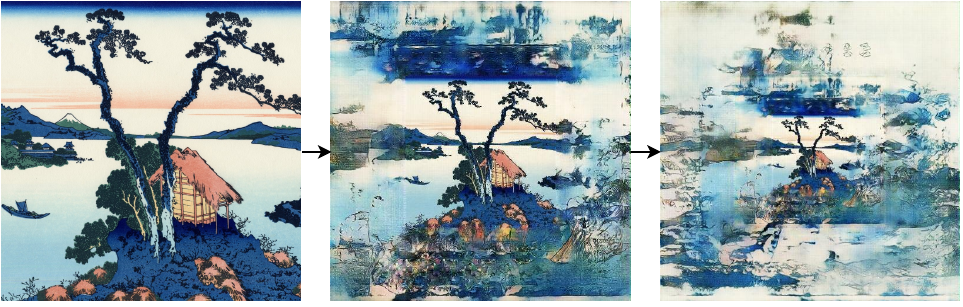}
    \caption{Two generations of continuation for Lake Suwa in the Shinano province, Katsushika Hokusai (circa 1829-1833).}
    \label{fig:ukiyo-continue-4}
\end{figure}
The model confusion that was noted previously due to the presence of the sky gradient is also observed in the continuation of Lake Suwa in the Shinano province by Katsushika Hokusai in Figure \ref{fig:ukiyo-continue-4}. Again, this has been formed into a mountain range and clouds, with the exception that in this image, the rooftops of a traditional Japanese building have been produced towards the top of the final image. This is likely due to the prominence of the depictions of temples in traditional Japanese art~\cite{sparavigna2016japanese}. The body of water has been extended, with new landmasses produced to populate it. The horizon has also been extended, with new mountains added, note the mountain on the left of the second image is generated to comes down from the peak but does not mirror the observed side. More colourful natural features are generated towards the closer areas of the image, before the second run generates a coastline with waves inspired by Hokusai's other works in the dataset. Waves similar to those produced are prominent in Hokusai's work~\cite{yura1967pedigree}, but the original image in particular did not contain any prior to their generation. 
\begin{figure}
    \centering
    \includegraphics{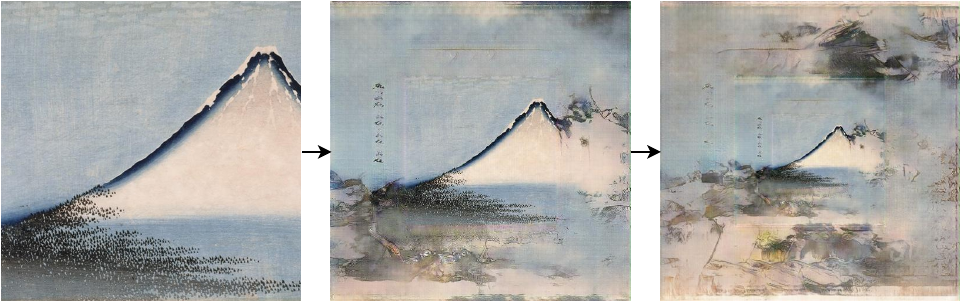}
    \caption{Two generations of continuation for Fuji Blue, Katsushika Hokusai (circa 1829-1833).}
    \label{fig:ukiyo-continue-5}
\end{figure}
Figure \ref{fig:ukiyo-continue-5} details the continuation of Hokusai's Fuji Blue. First and foremost, it can be noted that Mount Fuji, rather than being completed, is instead extended into a mountain range. Most interestingly in this image, the volumetric effect of the fog towards the bottom of the image is also extended, being filtered over the other features that the generator implements. As well as the addition of a grey cloud in the sky, the generator has also produced patterns which seems to be reminiscent of a village and natural features such as hills. 

\subsection{Abstract}
\begin{figure}
    \centering
    \includegraphics[scale=0.65]{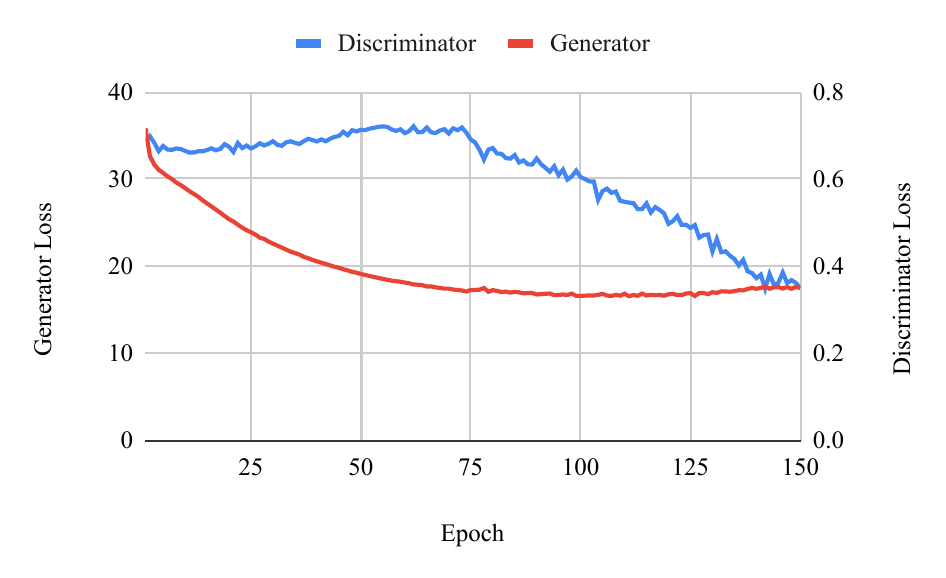}
    \caption{Observed Discriminator and Generator losses for the abstract art inpainting GAN.}
    \label{fig:abstract-disc-gen}
\end{figure}
\begin{figure}
    \centering
    \includegraphics[scale=0.65]{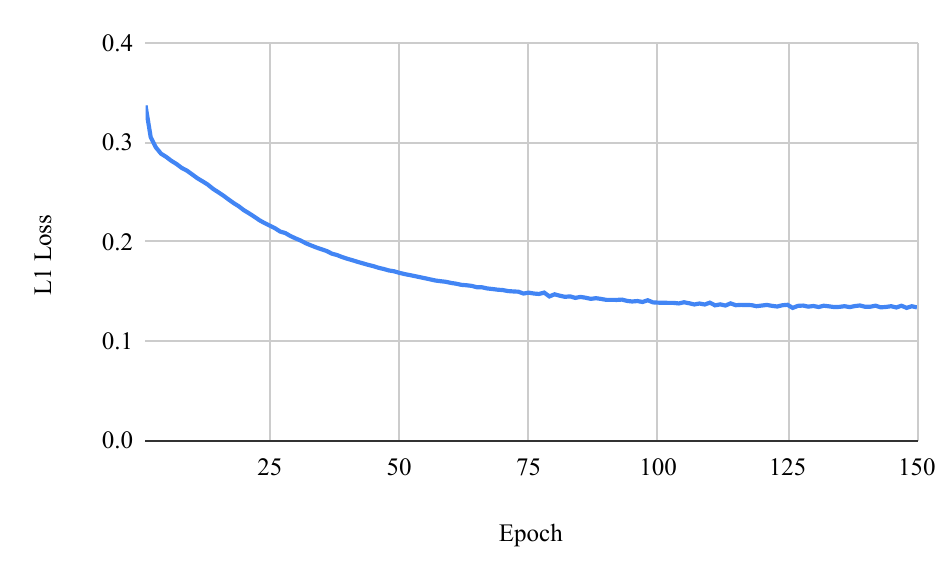}
    \caption{Observed generator L1 loss for abstract art inpainting.}
    \label{fig:abstract-l1}
\end{figure}
Figure \ref{fig:abstract-disc-gen} shows the particularly interesting behaviours of the discriminator and generator for the abstract process. Given the nature of abstract art, and the abstract outputs of a GAN's generator, it can be observed that classifying fake abstract art was particularly difficult for the discriminator. Whereas the generator improved rapidly, the discriminator trained for 75 epochs before being able to provide a challenge for the generator. Figure \ref{fig:abstract-l1} shows the generator's L1 loss, which, similarly to the other experiments, gradually decreased before stabilising.

\begin{figure}
    \centering
    \includegraphics{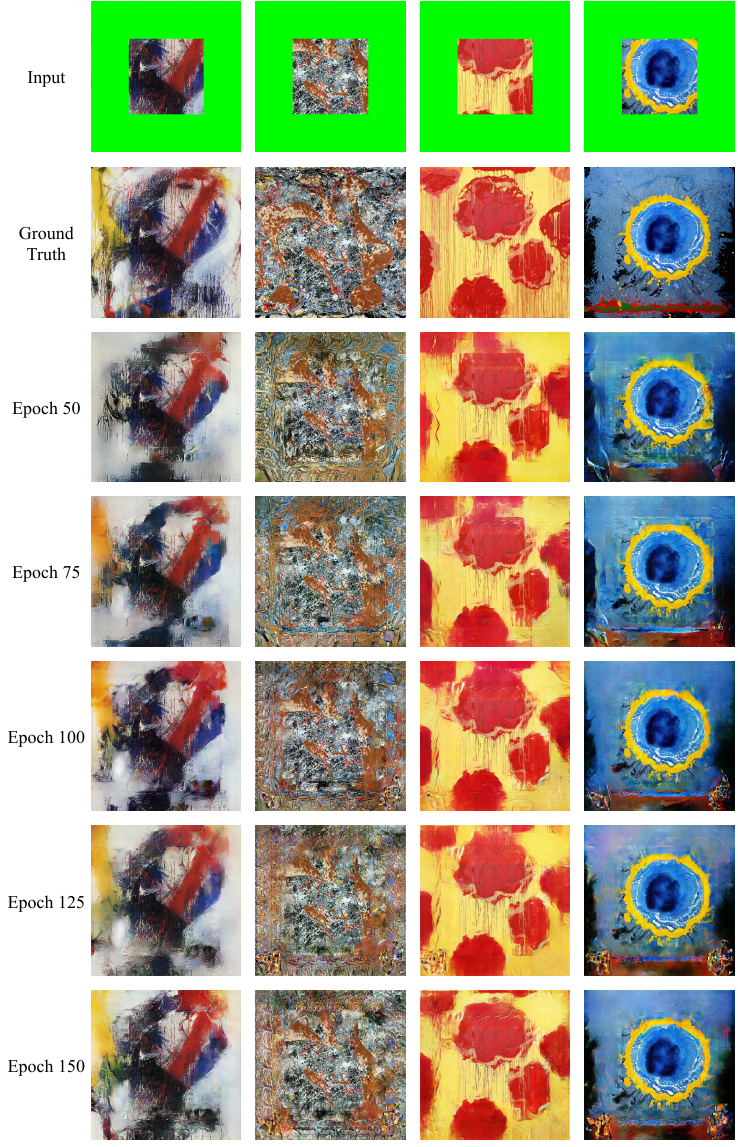}
    \caption{Examples of generator output for abstract art at selected epochs. }
    \label{fig:abstract-examples}
\end{figure}
Examples of training outputs can be observed in Figure \ref{fig:abstract-examples}. From left to right, the images shown for example are:
\begin{itemize}
    \item Excalibur, Norman Bluhm (1960)
    \item Number 7 C, Jackson Pollock (1949)
    \item Untitled, Cy Twombly (2007)
    \item Moon's Milk, John Hoyland (2009)
\end{itemize}
One of the more interesting things to note from these examples are the generator outputs for Untitled (3$^{rd}$ column) and Moon's Milk (4$^{th}$ column); note the form of the circles in these images is replicated early on, an impressive ability both due to the accuracy of the output at early stages as well as the nature of convolutions being square. It can also be observed that style such as running and spotted paint are produced in addition to realistic colourisation. 

\subsubsection{Continuation of abstract art}
\begin{figure}
    \centering
    \includegraphics{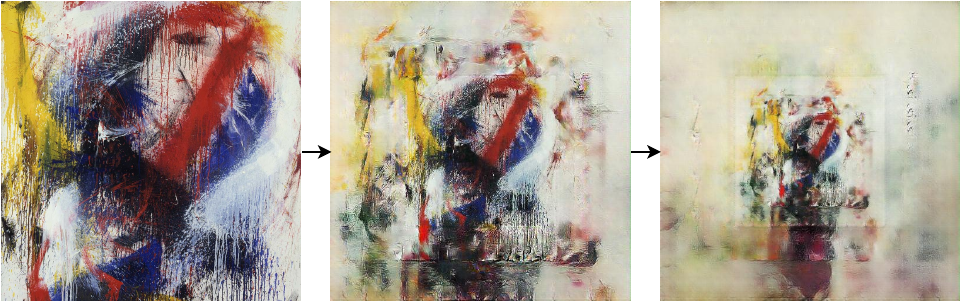}
    \caption{Two generations of continuation for Excalibur, Norman Bluhm (1960).}
    \label{fig:abstract-continue-1}
\end{figure}
Figure \ref{fig:abstract-continue-1} shows the continuation of Norman Bluhm's Excalibur for two generations. Although the first generation shows a continuation of paint, for example the yellow shape towards the left of the image being extended, as well as colours being mixed in a similar fashion, this does not seem to occur as much in the second run. In the second run, on the other hand, the canvas is extended with some sense of colours but far less prominently than in the original image or the first generation of continuation. 
\begin{figure}
    \centering
    \includegraphics{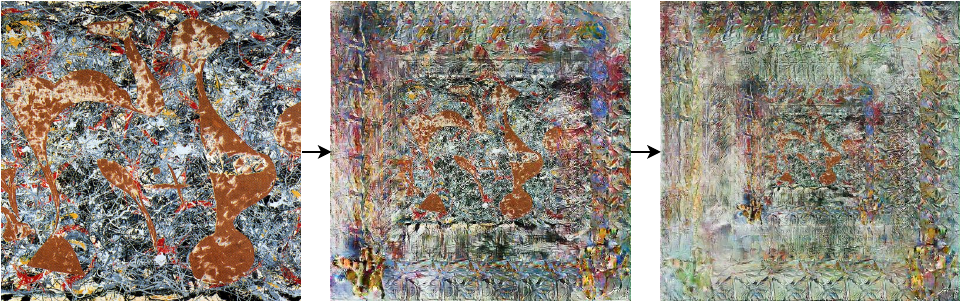}
    \caption{Two generations of continuation for Number 7 C, Jackson Pollock (1949).}
    \label{fig:abstract-continue-2}
\end{figure}
In Figure \ref{fig:abstract-continue-2}, the continuation of Jackson Pollock's Number 7 C can be observed. Once resized, the generator continues this now smaller pattern of paint throughout, but seems to implement a more unrealistic texture. This can be seen again in the final image but to a more prominent effect since the resizing has caused the spots of paint to become only several pixels large. 
\begin{figure}
    \centering
    \includegraphics{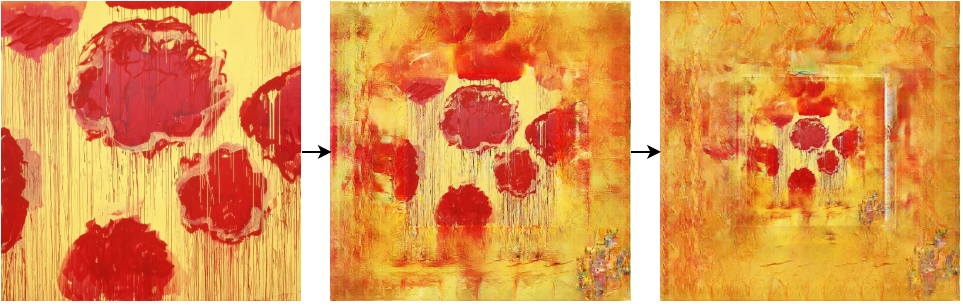}
    \caption{Two generations of continuation for Untitled, Cy Twombly (2007).}
    \label{fig:abstract-continue-3}
\end{figure}
Cy Twombly's Untitled is continued by the generator in Figure \ref{fig:abstract-continue-3}. In the first instance, the peony blossoms are continued but with less form than the original, being blended with the background to more of a degree. This continues throughout, with the added data in the second generation being much more generalised, although the pattern of running red paint on the yellow background is produced with considerable clarity.
\begin{figure}
    \centering
    \includegraphics{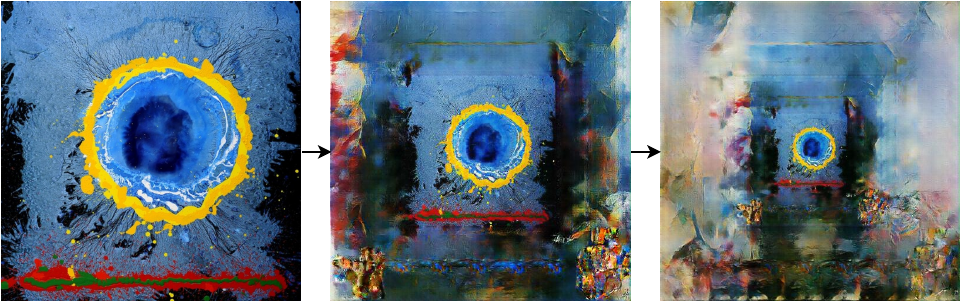}
    \caption{Two generations of continuation for Moon's Milk, John Hoyland (2009).}
    \label{fig:abstract-continue-4}
\end{figure}
In Figure \ref{fig:abstract-continue-4}, Moon's Milk by John Hoyland is continued by the generator. During both generations, the edges are extended to a certain extent, before new colours and patterns are introduced. In particular, the red and white paint towards the left of the generated images seem to be inspired by the shape at the bottom of the original image. Interestingly, the final generation seems to be fading to a more natural canvas colour, likely given that smaller images within the centre of the canvas, which is not completely painted are present within the dataset. This is particularly interesting since this is a feature not present in the original image, but becomes a behaviour invoked of the generator once the original sharper image has become relatively small. 
\begin{figure}
    \centering
    \includegraphics{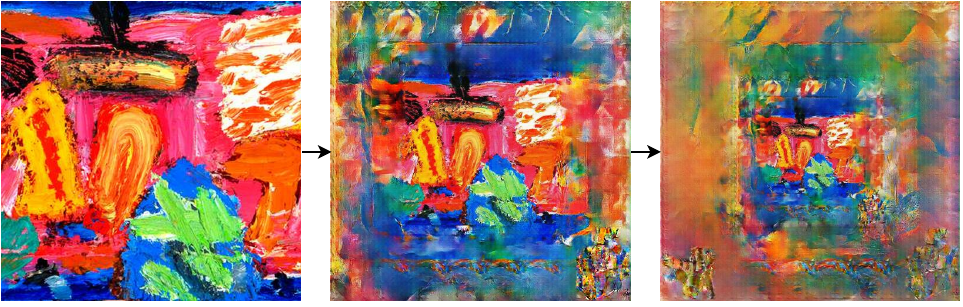}
    \caption{Two generations of continuation for Jarrow Wine, Basil Beattie (1985).}
    \label{fig:abstract-continue-5}
\end{figure}
As could be observed prior, the second generation of abstract art seems to lead to a loss of detail. This effect can also be seen in the continuation of Basil Beattie's Jarrow Wine, observed in Figure \ref{fig:abstract-continue-5}. Although this occurs, the second generation is particularly impressive, since the patterns are not only continued but created too. 

\subsection{Checkerboarding Artifacts}
As can be observed earlier in Figure \ref{fig:landscape-continue-2}, the outputs of the GAN sometimes presented with checkerboard-pattern artifacts. This is an open issue within the State-of-the-Art and occurs often in related works\cite{radford2015unsupervised,salimans2016improved,donahue2016adversarial,dumoulin2016adversarially}. This is likely due to deconvolution overlap during the second half of the modified U-Net as proposed by \cite{odena2016deconvolution}. Though it is possible that weights can be learnt to overcome this, some of the results in this work still suffer from it, especially the generated landscape images. Furthermore, if Figure \ref{fig:FID-graph} is observed, FID stagnated much quicker than the other two genres considered. According to \cite{dong2015image}, future work may find success in overcoming deconvolutional overlap and the aforementioned quality issues by performing image processing-based resizing rather than having the neural network perform the task. 
\begin{figure}
    \centering
    \includegraphics{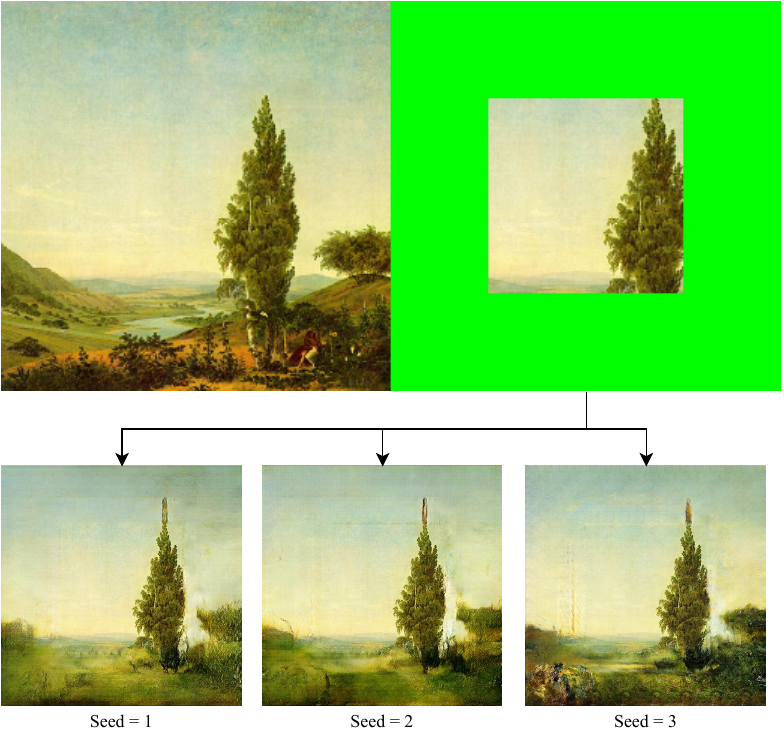}
    \caption{Outputs of GANs when the seed for random weight initialisation is changed (Caspar David Friedrich - Der Sommer, 1807).}
    \label{fig:weights-change}
\end{figure}
In Figure \ref{fig:weights-change}, it can be observed that some artifacts are caused by other factors in the learning process. Note that when the random weights are initialised with a seed of 3, more artifacts are present including some checkerboarding and a large entity generated in the bottom right corner of the image. Seeds of 1 and 2 are blurrier in their outputs, but seem more natural.

\section{Future Work and Conclusion}
\label{sec:conclusion}
The experiments in this work have shown that inpainting presents an exciting possibility as a process for generating artworks. Future work could involve data augmentation such as random noise, jitter, and cropping, as well as the exploration of other model hyperparameters. An extension to this project could also explore the possibility of a pipeline by way of implementing style transfer of the original image to the outputs of the generator. 
To finally conclude, this work has presented an image inpainting approach to generative art which diverges from the current state-of-the-art where images are synthesised either entirely generative by filtered random noise or inspired by the transfer of style. The experiments exploring landscapes, Ukiyo-e, and abstract art showed that, in many cases, features within the image were continued. Examples of this included the generation of new mountains and trees, as well as characters which resembled written text. In the abstract pieces, features such as spotted and running paint were also produced by the generator. 

Given that other approaches for inpainting for the specific case of generating AI art are yet to be presented, this work chose not to compare the models directly to the State-of-the-Art. This was due to the fact that it would be an unfair comparison, since the other models would be designed for a specific use-case that is different to the one faced by this work. In the future, when more inpainting methods for the synthesis of paintings are presented, then a more direct comparison would be possible. Training a GAN on modern computational hardware is computationally expensive and requires a relatively large amount of memory due to the use of both convolutional and deconvolutional operations\cite{shrivastava2021survey}. It is for this reason that hyperparameter optimisation is difficult, and impossible for high resolution images without access to a supercomputer e.g., StyleGan2's 25k 1024px image configuration would take over 35 days on a single GPU but 5 days on 8 GPUs\cite{karras2020analyzing}. In future, with either new hardware or access to such a system, models could be optimised in terms of their hyperparameters to explore how much of an effect they have and if they can be improved as such. In future, other applications of the approach could be explored. For example, the reconstruction of large entities such as photographs and smaller ones such as handwriting. The examples in this work showed that features such as trees and even whole villages could be constructed by the GAN, but also that features such as characters inspired by Japanese handwriting could also be synthesised. 

\bibliographystyle{IEEEtran}
\bibliography{bibliography}

\end{document}